\documentclass[runningheads]{llncs}
\usepackage[heightadjust=all]{floatrow}
\usepackage{graphicx}

\usepackage{tikz}
\usepackage{comment}
\usepackage{amsmath,amssymb} 
\usepackage{color}

\usepackage[accsupp]{axessibility}  
\usepackage{bm}
\usepackage[numbers,sort,compress]{natbib}
\usepackage{booktabs}
\usepackage{multirow}
\usepackage{subcaption}
\usepackage{xcolor}
\usepackage{capt-of}
\usepackage{algorithm}
\usepackage{algpseudocode}
\usepackage{wrapfig}
\usepackage{adjustbox}

\newcommand{\figref}[1]{Fig.~\ref{#1}}
\newcommand{\secref}[1]{Section~\ref{#1}}
\newcommand{\algoref}[1]{Algorithm~\ref{#1}}
\newcommand{\eqnref}[1]{Eq.~\eqref{#1}}
\newcommand{\tabref}[1]{Table~\ref{#1}}

\DeclareMathOperator*{\argmin}{arg\,min~}

\makeatletter
\DeclareRobustCommand\onedot{\futurelet\@let@token\@onedot}
\def\@onedot{\ifx\@let@token.\else.\null\fi\xspace}

\makeatother

\newcommand{\boldparagraph}[1]{\vspace{0.2cm}\noindent{\bf #1:} }

\def\ind{\hspace*{\algorithmicindent}}

\begin{document}

\pagestyle{headings}
\mainmatter
\def\ECCVSubNumber{5325}  

\title{Improving the Intra-class Long-tail in 3D Detection via Rare Example Mining} 

\titlerunning{Rare Example Mining for 3D Detection}
%
\author{Chiyu Max Jiang \and
Mahyar Najibi \and
Charles R. Qi \and
\\
Yin Zhou \and
Dragomir Anguelov}
\authorrunning{C. Jiang et al.}
%
\institute{Waymo LLC., Mountain View CA 94043, USA \\
\email{\{maxjiang,najibi,rqi,yinzhou,dragomir\}@waymo.com}}
\maketitle

\setlength{\abovedisplayskip}{0.25\abovedisplayskip}
\setlength{\belowdisplayskip}{0.25\belowdisplayskip}
\setlength{\abovecaptionskip}{0.5\abovecaptionskip}
\setlength{\belowcaptionskip}{0.25\belowcaptionskip}

\setlength{\textfloatsep}{10pt plus 1.0pt minus 2.0pt}
\setlength{\floatsep}{12pt plus 1.0pt minus 1.0pt}
\newfloatcommand{capbtabbox}{table}[][\FBwidth]
\newfloatcommand{capbalgbox}{algorithm}[][\FBwidth]

\begin{abstract}
Continued improvements in deep learning architectures have steadily advanced the overall performance of 3D object detectors to levels on par with humans for certain tasks and datasets, where the overall performance is mostly driven by common examples. However, even the best performing models suffer from the most naive mistakes when it comes to rare examples that do not appear frequently in the training data, such as vehicles with irregular geometries. Most studies in the long-tail literature focus on class-imbalanced classification problems with known imbalanced label counts per class, but they are not directly applicable to the intra-class long-tail examples in problems with large intra-class variations such as 3D object detection, where instances with the same class label can have drastically varied properties such as shapes and sizes. Other works propose to mitigate this problem using active learning based on the criteria of uncertainty, difficulty, or diversity. In this study, we identify a new conceptual dimension - rareness - to mine new data for improving the long-tail performance of models. We show that rareness, as opposed to difficulty, is the key to data-centric improvements for 3D detectors, since rareness is the result of a lack in data support while difficulty is related to the fundamental ambiguity in the problem. We propose a general and effective method to identify the rareness of objects based on density estimation in the feature space using flow models, and propose a principled cost-aware formulation for mining rare object tracks, which improves overall model performance, but more importantly - significantly improves the performance for rare objects (by 30.97\%).
\end{abstract}

\section{Introduction}
\label{sec:intro}

\begin{figure}[t]
\begin{floatrow}
\ffigbox[.46\textwidth]{%
    \centering
    \includegraphics[width=\linewidth]{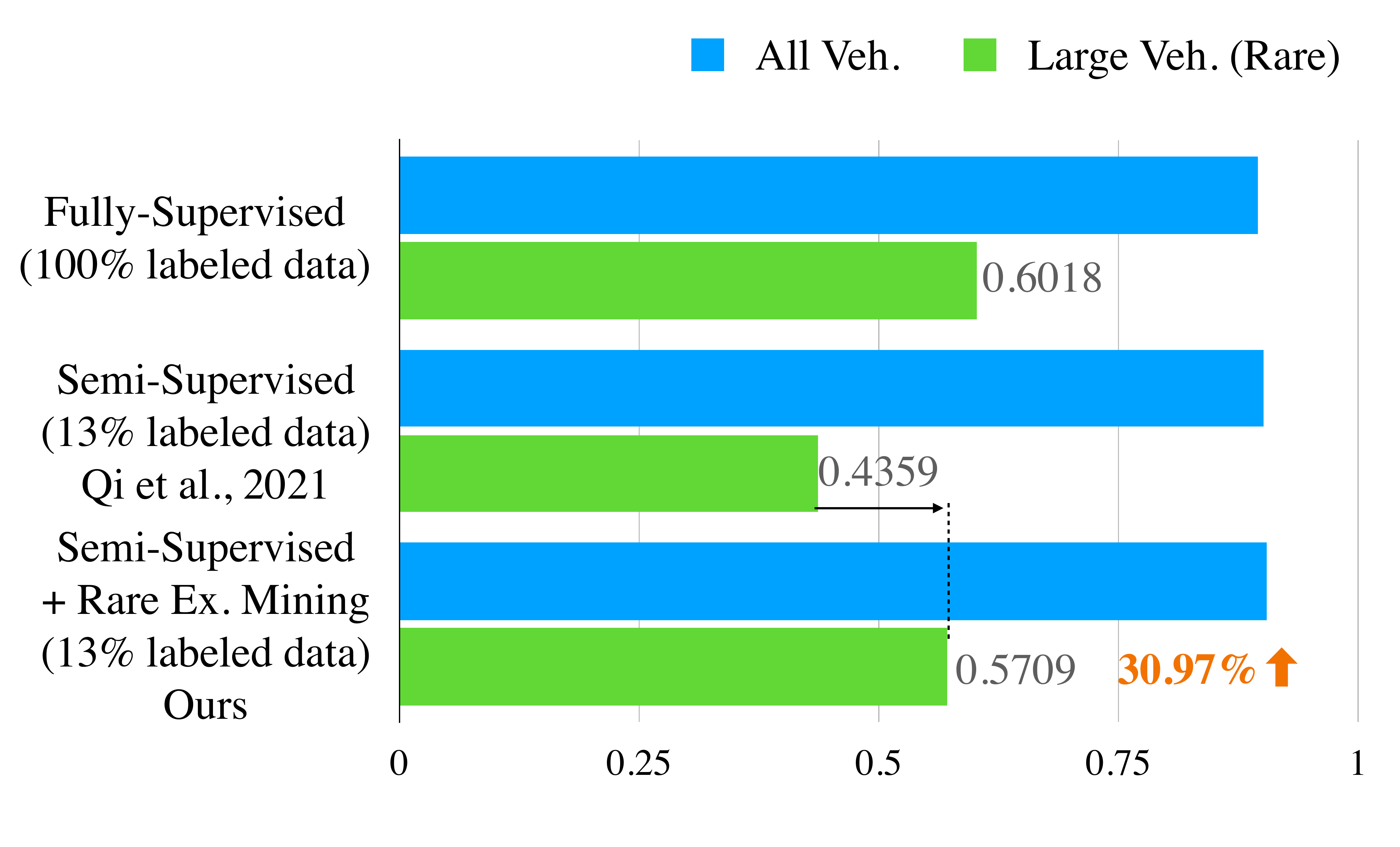}
}{
    \caption{Vehicle 3D object detection Average Precision (AP) on the Waymo Open Dataset with fully-/semi-supervised learning. While standard semi-supervised learning (with a strong auto labeling teacher model \cite{qi2021offboard}) can achieve on par results with fully supervised method on the common cases, the performance gap on rare objects (e.g. large vehicles) is significant (60.18 v.s. 43.59). Our method is able to close this gap using rare example mining.}
    \label{fig:main_result}
}
\ffigbox[.51\textwidth]{%
    \includegraphics[width=\linewidth]{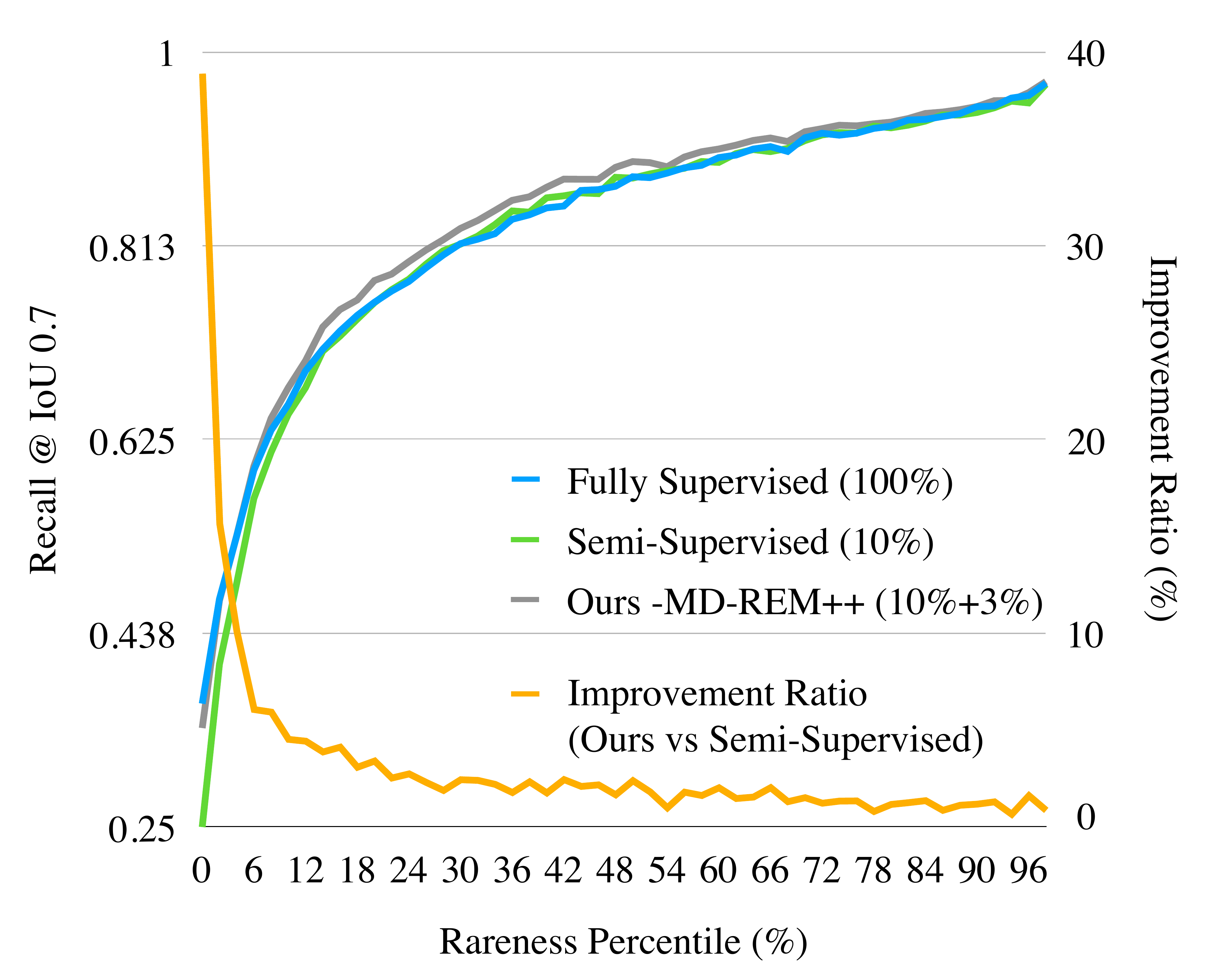}
}{
    \caption{Correlation between inferred rareness percentile (lower is more rare) and model performance for subsets of ground truth, indicated by recall. In all models (from fully-supervised to semi-supervised), model performance is strongly correlated to the rareness measure obtained from the log probability inferred by the flow model. By mining a mere 3\% of remaining data, our model significantly improves upon the semi-supervised detector, with big gains in the rare intra-class long-tail.}
     \label{fig:rare_vs_recall}
}
\end{floatrow}
\end{figure}

Long-tail learning is a challenging yet important topic in applied machine learning, particularly for safety-critical applications such as autonomous driving or medical diagnostics. However, even though imbalanced classification problems have been heavily studied in the literature, we have limited tools in defining, identifying, and improving on intra-class rare instances, such as irregularly shaped vehicles or pedestrians in Halloween costumes, since they come from a diverse open set of anything but common objects. Inspired by Leo Tolstoy's famous quote, we observe: ``Common objects are all alike; Every rare object is rare in its own way''.

We refer to the spectrum of such rare instances as the \emph{intra-class long-tail}, where we do not have the luxury of prespecified class-frequency-based rareness measurements. Objects of the intra-class long-tail can be of particular importance in 3D detection due to its safety relevance. While overall performance for modern 3D detectors can be quite high, we note that even fully supervised models perform significantly worse on rare subsets of the data, such as large vehicles (\figref{fig:main_result}). The problem is exacerbated by semi-supervised learning, a popular and cost-efficient approach to quickly scale models on larger datasets where average model performance have been shown to be on par with fully-supervised counterparts using a fraction of the labeled data.

\begin{figure*}[t!]
    \centering
    \includegraphics[width=\textwidth]{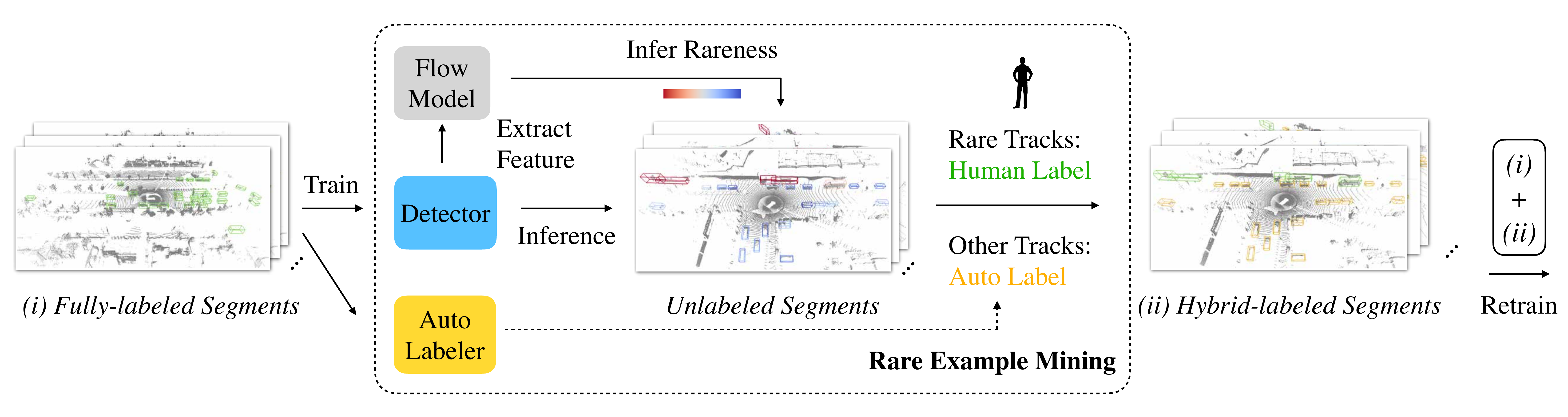}
    \caption{Overview of the Rare Example Mining (REM) pipeline. Our detector, bootstrap-trained on a smaller pool of fully labeled segments, extracts features for a flow model to infer the log probability of every detected instance, which is a strong indicator of rareness. The rare tracks in the unlabeled segments are sent for human labeling while all remaining tracks are labeled using an offboard auto-labeler. The combined datasets is then used for retraining the detector, resulting in an overall performance boost, particularly on rare examples.}
    \label{fig:pipeline}
\end{figure*}

Several challenges make it difficult for targeted improvement on the intra-class long-tail for 3D detection. First, as box regression is an important aspect of object detection, conventional long-tail learning approaches utilizing class frequencies, or active learning approaches utilizing entropy or margin uncertainties that depend on classification output distributions are not applicable. Second, since labeling cost given a run segment is proportional to the number of labeled instance tracks, not frames, we require a more granular mining approach that gracefully handles missing labels for objects in the scene. Last but not least, unlike long-tail problems for imbalanced classification tasks, it is challenging to define which examples belongs to the intra-class long-tail, which leads to difficulty in evaluating and mining additional data to improve the long-tail performance of these models.

In light of these challenges, we propose a generalizable yet effective way to measure and define rareness as the density of instances in the latent feature space. We discover that normalizing flow models are highly effective for feature density estimation and robust for anomaly detection, contrary to negative results on anomaly detection using normalizing flows directly on high dimensional image inputs, as reported by prior work \cite{nalisnick2018deep}. We present a cost-aware formulation for track-level data-mining and active learning using the rareness criteria, as 3D object labeling cost is often proportional to the number of unique tracks in each run segment. We do this in conjunction with a powerful offboard 3D auto-labeler \cite{qi2021offboard,yang2021auto4d} for filling in missing data, and show stronger model improvement compared to difficulty, uncertainty, or heuristics based active learning baselines, particularly for objects in the tail distributions.

Furthermore, we investigate rareness as a novel data-mining criterion, in relation to the conventional uncertainty or error-based mining methods. Though models tend to perform poorly on either rare or hard examples, we note a clear distinction between the concept of rare versus hard. In this discussion, ``rare'' maps to epistemic uncertainty (reducible error) where the model is uncertain due to a lack of data support in the training set, while ``hard'' maps to aleatoric uncertainty (irreducible error), where the model is uncertain due to the fundamental ambiguity and uncertainty of the given problem, for example, if the target object is heavily occluded. We further illustrate that while conventional uncertainty estimates (such as ensembling methods) will uncover both hard and rare objects, filtering out hard examples will result in a significantly higher concentration of rare examples which significantly improves active learning performance, underscoring the importance of rare examples in active learning.

In summary, the main contributions of this work are:
\begin{itemize}
    \item We identify rareness as a novel criterion for data mining and active learning, for improving model performance for problems with large intra-class variations such as 3D detection.
    \item We propose an effective way of identifying rare objects by estimating latent feature densities using a flow model, and demonstrate a strong correlation between estimated log probabilities, known rare subcategories, and model performance.
    \item We propose a fine-grained, cost-aware, track level mining methodology for 3D detection that utilizes a powerful offboard 3D auto-labeler for annotating unlabeled objects in partially labeled frames, resulting in a strong performance boost (30.97\%) on intra-class long-tail subcategories compared to convetional semi-supervised baselines.
\end{itemize}

\section{Related Work}
\label{sec:related_work}

\boldparagraph{Long-tail visual recognition}%
Long-tail is conventionally defined as an imbalance in a multinomial distribution between various different class labels, either in the image classification context \cite{liu2019large,jamal2020rethinking,kang2019decoupling,wang2017learning,cui2019class,kim2020m2m,zhong2021improving,zhao2021improving}, dense segmentation problems \cite{wu2020forest,hsieh2021droploss,wang2019classification,gupta2019lvis,zang2021fasa,wang2020seesaw}, or between foreground / background labels in object detection problems \cite{zhang2021simple,tan2020equalization,tan2020equalizationv2,lin2017focal,li2020overcoming}. Existing approaches for addressing class-imbalanced problems include resampling (oversampling tail classes or head classes), reweighitng (using inverse class frequency, effective number of samples \cite{cui2019class}),  novel loss function design \cite{lin2017focal,tan2020equalization,tan2020equalizationv2,wang2020seesaw,abdelkarim2020long,zheng2018ring}, meta learning for head-to-tail knowlege transfer \cite{wang2017learning,kim2020m2m,liu2021gistnet,chu2020feature}, distillation \cite{li2021self,xiang2020learning} and mixture of experts \cite{wang2020long}.

However, there is little work targeting improvements for the intra-class long-tail in datasets with inherently large intra-class variations, or for regression problems. %
\citet{zhu2014capturing} studies the long-tail problem for subcategories, but assumes given subcategory labels. \citet{dong2017class} studies imbalance between fine-grained attribute labels in clothing or facial datasets. To the best of our knowledge, our work is among the first to address the intra-class long-tail in 3D object detection.

\boldparagraph{Active learning}%
In this work we mainly address pool-based active learning \cite{settles2009active}, where we assume an existing smaller pool of fully-labeled data along with a larger pool of unlabeled data, from which we actively select samples for human labeling. Existing active learning methods mainly fall under two categories, uncertainty-based and diversity-based methods. Uncertainty-based methods select new labeling targets based on criteria such as ensemble variance \cite{beluch2018power} or classification output distribution such as entropy, margin or confidence \cite{holub2008entropy,joshi2009multi,qi2008two,gal2017deep,choi2021active,harakeh2020bayesod} in the case of classification outputs. More similar to our approach are diversity-based approaches, that aim at balancing the distribution of training data while mining from the unlabeled pool \cite{gudovskiy2020deep,guo2010active,nguyen2004active,sener2017active}. \citet{gudovskiy2020deep} further targets unbalanced datasets. However, these methods are developed for classification problems and are not directly applicable to the intra-class long-tail for detection tasks. Similar to our approach, \citet{sinha2019variational} proposes to learn data distributions in the latent space, though they employ a discriminator in a variational setting that does not directly estimate the density of each data sample. \citet{segal2021just} investigated fine-grained active learning in the context self-driving vehicles using region-based selection with a focus on joint perception and prediction. Similar to our approach, \citet{elezi2022not} uses auto-labeling to improve active learning performances for 2D detection tasks.

\boldparagraph{Flow models}
Normalizing flow models are a class of generative models that can approximate probability distributions and efficiently and exactly estimate the densities of high dimensional data \cite{dinh2014nice,rezende2015variational,dinh2016density,kingma2018glow,chen2018neural,grathwohl2018ffjord,kobyzev2020normalizing}. Various studies have reported unsuccessful attempts at using density estimations estimated by normalizing flows for detecting out-of-distribution data by directly learning to map from the high dimensional pixel space of images to the latent space of flow models \cite{nalisnick2018deep,choi2018waic,zhang2021understanding}, assigning higher probability to out-of-distribution data. However, similar to our finding, \citet{kirichenko2020normalizing} find that the issue can be easily mitigated by training a flow model on the features extracted by a pretrained model such as an EfficientNet pretrained on ImageNet \cite{deng2009imagenet}, rather than directly learning on the input pixel space. This allows the model to better measure density in a semantically relevant space. We are among the first to use densities estimated by normalizing flows for identifying long-tail examples.

\section{Methods}
In this section, we present a general and effective method for mining rare examples based on density estimations from the data, which we refer to as data-centric rare example mining (REM). To offer further insights to rareness in relation to difficulty, we propose another conceptually simple yet effective method for mining rare examples by simply filtering out hard examples from overall uncertain examples. In \secref{sec:exp_active}, we show that combining both approaches can further improve long-tail performance. Last but not least, we propose a cost-aware, fine-grained track-level active learning method that aggregates per-track rareness as a selection criteria for requesting human annotation, and utilize a powerful offboard 3D auto-labeler for populating unmined, unlabeled tracks to maximize the utility of all data when retraining the model.

\subsection{Rare Example Mining}

\subsubsection{Data-centric Rare Example Mining (D-REM)}

The main intuition behind data-centric REM is that we measure the density of every sample in a learned feature embedding space as an indicator for rareness. 

The full data-centric REM workflow (see \figref{fig:pipeline}) consists of the following steps. First, we pretrain the detection model on an existing source pool of fully-labeled data that might be underrepresenting long-tail examples. Second, we use the pretrained task model to run inference over the source pool along with a large unlabeled pool of data, and extract per-instance raw feature vectors via Region-of-interest (ROI) pooling, followed by PCA dimensionality reduction and normalization. We then train a normalizing flow model over the feature vectors to estimate per-instance rareness (negative log probability) for data mining. 

\begin{figure}[t!]
     \centering
     \includegraphics[width=\linewidth]{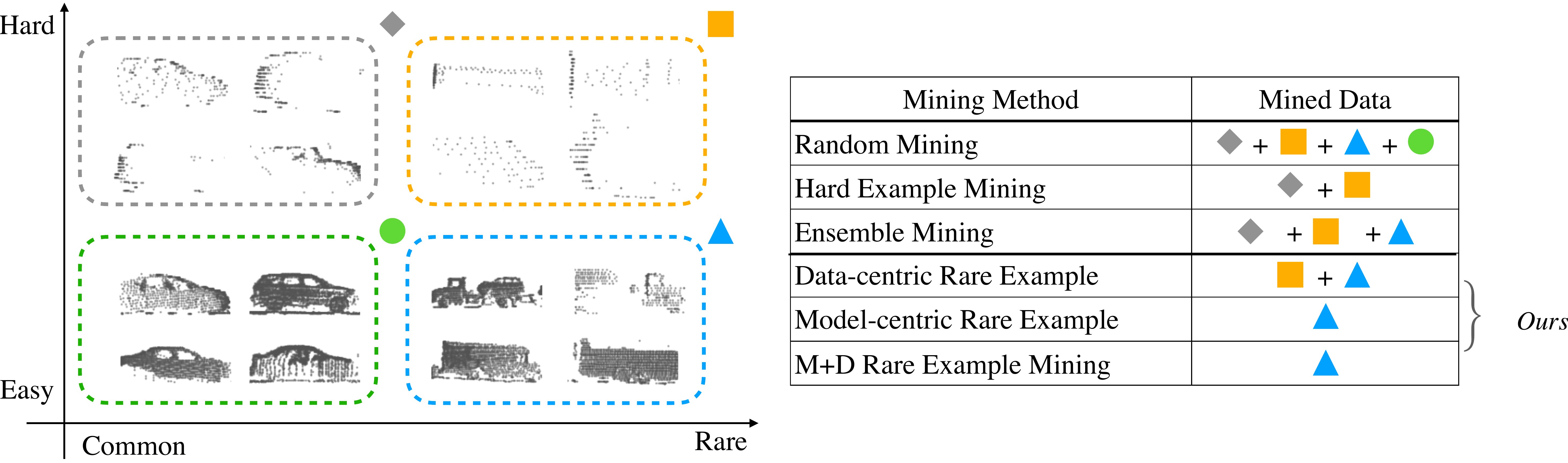}
     \caption{Hard (aleatoric uncertainty) is a fundamentally different dimension compared to rare (epistemic uncertainty). Our REM method directly targets rare subsets of the data. Our Data-centric REM method directly estimates rareness based on inferred probabilities by a normalizing flow model trained on learned feature vectors, while our Model-centric REM method performs hard example filtering on top of generic uncertain objects mined by the ensemble mining approach. We further combine the two approaches (MD-REM) by performing hard example filtering on top of D-REM to increase easy-rare examples.}
     \label{fig:rare_vs_hard}
\end{figure}

\boldparagraph{Object Feature Extraction} As previously mentioned, one major difference between our proposed approach for estimating rare examples, compared with earlier works in the literature that were not successful in using normalizing flow for out-of-distribution detection \cite{nalisnick2018deep,choi2018waic,zhang2021understanding}, is that we propose to estimate the probability density of each example in the latent feature space of pretrained models to leverage the semantic similarity between objects for distinguishing rare instances. As observed by \citet{kirichenko2020normalizing}, normalizing flow directly trained on high dimensional raw input features tend to focus more on local pixel correlations rather than semantics as it doesn't leverage high-level embeddings.

We extract per-object feature embeddings from the final Birds-Eye-View (BEV) feature map of a 3D object detector via region of interest (ROI) max-pooling \cite{girshick2014rich} by cropping the feature map with the prediction boxes.  We mainly apply this for our implementation of the state-of-the-art MVF \cite{zhou2020end,qi2021offboard} 3D detector, though the process is generally applicable to majority of detectors that produce intermediate feature maps \cite{lang2019pointpillars,meyer2019lasernet,sun2021rsn}.

We further perform principal component analysis (PCA) for dimensionality reduction for improved computational efficiency, followed by normalization on the set of raw feature vectors $X_{\text{roi}}\in \mathbb{R}^{n\times d}$ obtained via ROI pooling

\begin{align}
    X_{\text{pca}} &= (X_{\text{roi}} - mean(X_{\text{roi}}))W_{\text{pca}}^{T}  \\
    X_{\text{norm}} &= X_{\text{pca}}~/~std(X_{\text{pca}})
\end{align}
where $W_{\text{pca}}\in\mathbb{R}^{k\times d}$ is a weight matrix consisting of the top-$k$ PCA components, $\text{mean}(\cdot): \mathbb{R}^{n\times d} \mapsto \mathbb{R}^{d}$, $std(\cdot): \mathbb{R}^{n\times d} \mapsto \mathbb{R}^{d}$ are the mean and standard deviation operators along the first dimension.

In summary, the training dataset for our flow model consists of normalized feature vectors after PCA-transformation obtained via ROI max-pooling the final feature map of 3D detectors using predicted bounding boxes.
\begin{equation}
    \mathcal{D}_{x} = \{X_{\text{norm}}[i], \forall i\in[0, n)\}
\end{equation}

\boldparagraph{Rareness Estimation Using Normalizing Flow} We use the continuous normalizing flow models for directly estimating the log probability of each example represented as a feature vector $\bm{x}$. We present a quick review of normalizing flows below.

Typical normalizing flow models \cite{kingma2018glow} consist of two main components: a base distribution $p(\bm{z})$, and a learned invertible function $f_{\bm{\theta}}(\bm{x})$, also known as a bijector, where $\bm{\theta}$ are the learnable parameters of the bijector, $f_{\bm{\theta}}(\bm{x})$ is the forward method and $f_{\bm{\theta}}^{-1}(\bm{x})$ is the inverse method. The base distribution is generally chosen to be an analytically tractable distribution whose probability density function (PDF) can be easily computed, such as a spherical multivariate Gaussian distribution, where $p(\bm{z}) = \mathcal{N}(\bm{z};0,\bm{I})$. A learnable bijector function can take many forms, popular choices include masked scale and shift functions such as RealNVP \cite{dinh2016density,kingma2018glow} or continuous bijectors utilizing learned ordinary differential equation (ODE) dynamics \cite{chen2018neural,grathwohl2018ffjord}.

The use of normalizing flows as generative models has been heavily studied in the literature \cite{kingma2018glow}, where new in-distribution samples can be generated via passing a randomly sampled latent vector through the forward bijector:
\begin{equation}
    \bm{x} = f_{\bm{\theta}}(\bm{z}),\quad \text{where}~\bm{z}\sim p(z)
\end{equation}

However, in this work, we are more interested in using normalizing flows for estimating the exact probabilities of each data example. The latent variable corresponding to a data example can be inferred via $\bm{z} = f_{\bm{\theta}}(\bm{x})$. Under a change-of-variables formula, the log probability of a data sample can be estimated as:
\begin{align}
    \log p_{\theta}(\bm{x}) &= \log p(f_{\bm{\theta}}(\bm{x})) + \log |\det (df_{\bm{\theta}}(\bm{x})/d\bm{x})| \\
    &= \log p(\bm{z}) + \log |\det (d\bm{z}/d\bm{x})|
\end{align}
The first term, $\log p(\bm{z})$, can be efficiently computed from the PDF of the base distribution, whereas the computation of the log determinant of the Jacobian: $\log |\det (df_{\bm{\theta}}(\bm{x})/d\bm{x})|$ vary based on the bijector type.

The training process can be described as a simple maximization of the expected log probability of the data (or equivalently minimization of the expected negative log likelihood of the parameters) from the training data $\mathcal{D}_{\bm{x}}$ and can be learned via batch stochastic gradient descent:
\begin{equation}
    \argmin_{\theta} \mathbb{E}_{x\sim \mathcal{D}_{\bm{x}}} [-\log p_{\bm{\theta}}(\bm{x})]
\end{equation}
In our experiments, we choose the base distribution $p(\bm{z})$ to be a spherical multivariate Gaussian $\mathcal{N}(\bm{z};0,\bm{I})$, and we use the FFJORD \cite{grathwohl2018ffjord} bijector.

For the final rare example scoring function for the $i$-th object, $r_i$, we have:

\begin{equation}
    r_i = -\log p_{\bm{\theta}} (\bm{x_i}) \label{eqn:rem_data_scoring}
\end{equation}

\subsubsection{Model-centric Rare Example Mining (M-REM)}
We present an alternative model-centric formulation for REM that is conceptually easy and effective, yet illustrative of the dichotomy between rare and hard examples. Different from the data-centric REM perspective, model-centric REM leverages the divergence among an ensemble of detectors as a measurement of total uncertainty.

Different from methods that directly use ensemble divergence as a mining critera for active learning \cite{beluch2018power}, our key insight is that while ensemble divergence is a good measurement of the overall uncertainties for an instance, it could be either due to the problem being fundamentally difficult and ambiguous (i.e., hard), or due to the problem being uncommon and lack training support for the model (i.e., rare). In the case of 3D object detection, a leading reason for an object being physically hard to detect is occlusion and low number of LiDAR points from the object. Conceptually, adding more hard examples such as far-away and heavily occluded objects with very few visible LiDAR points would not be helpful, as these cases are fundamentally ambiguous and cannot be improved upon simply with increased data support.

A simple approach for obtaining rare examples, hence, is to filter out hard examples from the set of overall uncertain examples. In practice, a simple combination of two filters: (i) low number of LiDAR points per detection example, or (ii) a large distance between the detection example and the LiDAR source, proves to be surprisingly effective for improving model performance through data mining and active learning.

We implement model-centric REM as follows. Let $\mathcal{M} = \{M_{1}, M_{2}, \cdots, M_{N}\}$ be a set of $N$ independently trained detectors with identical architecture and training configurations, but different model initialization. Denote detection score for the $i$-th object by the $j$-th detector as $s_{i}^{j}$. $s_{i}^{j}$ is set to zero if there is a missed detection. The detection variance for the $i$-th object by the model ensemble $\mathcal{M}$ is defined as:
\begin{equation}
    v_i = \frac{1}{N} \sum_{j=1}^{N} (s_i^j - \frac{1}{N} \sum_{k=1}^{N} s_i^k)^2
    \label{eqn:ens_var}
\end{equation}

For hard example filtering, denote the number of LiDAR points within the $i$-th object as $p_i$, and the distance of the $i$-th object from the LiDAR source as $d_i$. A simple hard example filter function can be defined as:

\begin{equation}
    h_i = 1 \text{~if~} (p_i > \tilde{p})~\texttt{\&}~(d_i < \tilde{d}) \text{~else~} 0
\end{equation}

where $\tilde{p}, \tilde{d}$ are the respective point threshold and distance thresholds. In our experiments, we have $N = 5, \tilde{p} = 200, \tilde{d} = 50~(\text{meters})$.

The final rare example scoring function for the $i$-th object, $r_i$, can be given as:

\begin{equation}
    r_i = h_i * v_i \label{eqn:rem_model_scoring}
\end{equation}

\subsection{Track-level REM for Active Learning}\label{sec:track_rem}

\begin{figure}[t!]
\begin{floatrow}\BottomFloatBoxes
\ffigbox[.48\textwidth]{%
    \centering
    \includegraphics[width=\linewidth]{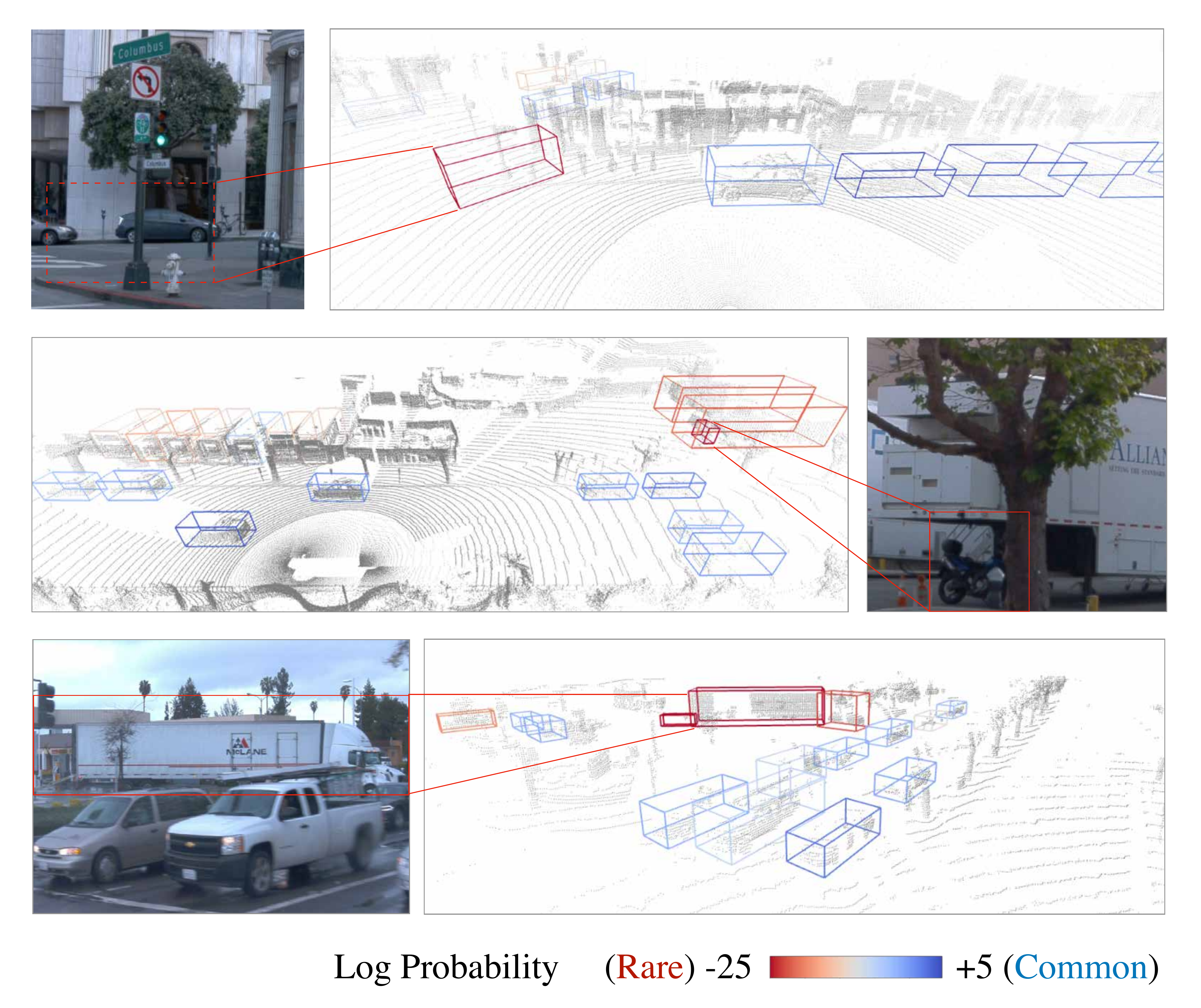}
}{
    \caption{Visualization of the rarest object tracks in the Waymo Open Dataset based on log probability inferred by the data-centric REM algorithm, where low log probability indicates rareness. The most rare instances include incorrectly labeled ground truth boxes, motorcycle underneath a trailer, and large vehicles.}
    \label{fig:viz_rare_common}
}
\capbalgbox[.48\textwidth]{
    \footnotesize
    \caption{Track-level REM}
    \label{alg:track_rem}
    \raggedright
    \textbf{Input} \\
    \ind Model detections $\mathcal{D}_b = b_1, b_2, \cdots, b_n$ (sort by descending rareness score) \\
    \ind Auto-labeled tracks $\mathcal{S}_{a} = \{T'_1, T'_2, \cdots, T'_m\}$ \\
    \ind Labeling budget $K$ \\
    \textbf{Output} \\
    \ind Fully labeled tracks $\mathcal{S} = \mathcal{S}_h \cup (\mathcal{S}_a - (\mathcal{S}_h \cap \mathcal{S}_a))$ \\
    \begin{algorithmic}[1]
    \Procedure{TrackMining}{}
    \State $\mathcal{S}_h \gets \{$\O$\}$
    \While{$|\mathcal{S}_h| < K$}
    \State $b \gets \mathcal{D}_b.pop(0)$
    \If{HumanCheckExists($b$)}
    \State $T = $HumanLabelTrackFromBox($b$)
    \State $\mathcal{S}_h.push(T)$
    \State $\mathcal{D}_b \gets$ DiscardIntersectingBoxes($\mathcal{D}_b, T$)
    \EndIf
    \EndWhile
    \State $\mathcal{S}_a \gets \mathcal{S}_a - (\mathcal{S}_h \cap \mathcal{S}_a)$ \\
    \Return $\mathcal{S} = \mathcal{S}_a \cup \mathcal{S}_h$
    \EndProcedure
    \end{algorithmic}
}{
}
\end{floatrow}
\end{figure}

To apply our REM method towards active learning as a principled way of collecting rare instances from a large unlabeled pool in a cost-effective manner, we propose a novel track-level mining and targeted annotation strategy in conjunction with a high-performance offboard 3D auto-labeler for infilling missing labels. We choose to mine at the track-level because labeling tools are optimized to label entire object tracks, which is cheaper than labeling per frame. Please refer to \figref{fig:pipeline} for an overview of the active learning pipeline and \algoref{alg:track_rem} for a detailed breakdown of the mining process.

First, starting with a labeling budget of $K$ tracks, we score each detected object from the unlabeled dataset using one of the rare example scoring functions above (Eq. (\ref{eqn:rem_data_scoring}, \ref{eqn:rem_model_scoring})). Starting from the detection object with the highest rareness score, we sequentially route each example to human labelers for labeling the entire track $T$ corresponding to the object and add the track to the set of mined and human-labeled tracks $\mathcal{S}_h$. Then all model detections that intersect with $T$ ($>0$ IoU) are removed. This procedure is iteratively performed until the number of tracks in $\mathcal{S}_h$ reaches the budget of $K$. All auto-labeled tracks $\mathcal{S}_a$ that intersect with $\mathcal{S}_h$ are removed, and the two sets of tracks are merged into a hybrid, fully-labeled dataset $\mathcal{S} = \mathcal{S}_a \cup \mathcal{S}_h$.

\section{Experiments}

We use the Waymo Open Dataset \cite{sun2020scalability} as the main dataset for our investigations due to its unparalleled diversity based on geographical coverage, compared with other camera+LiDAR datasets available \cite{geiger2013vision,caesar2020nuscenes}, as well as its large industry-level scale. The Waymo Open Dataset consists of 1150 scenes that span 20 seconds, recorded across a range of weather conditions in multiple cities.

In the experiments below, we seek to answer three questions: (1) Does model performance correlate with our rareness measurement for intra-class long-tail (\secref{sec:rem_analysis}), (2) Can our proposed rare example mining methodology successfully find and retrieve more rare examples (\secref{sec:rem_analysis}), and (3) Does adding rare data to our existing training data in an active learning setting improve overall model performance, in particular for the long-tail (\secref{sec:exp_active}).

\subsection{Rare Example Mining Analysis} \label{sec:rem_analysis}

In this section, we investigate the ability of the normalizing flow model in our data-centric REM method for detecting intra-class long-tail examples. 

\boldparagraph{Correlation: Rareness and Performance}
We investigate the correlation between the rareness metric (as indicated by low inferred log probability score on ground-truth labels), and the associated model performance on these examples, as measured by recall on GT examples grouped by rareness. We present the results in \figref{fig:rare_vs_recall}. All ground-truth examples are grouped by sorting along their inferred log probability (from an MVF and flow model trained on 100\% data) into 2\% bins. Recall metric for different experiments are computed for each bin. More details on our active learning experiment will be presented in \secref{sec:exp_active}.

We derive two main conclusions: (1) the performance for all models are strongly correlated with our proposed rareness measurement, indicating our flow probability-based estimation of rareness is highly effective. (2) Our proposed rare example mining method achieves significant performance improvement on rare examples compared to the original semi-supervised baseline using a small fraction of additional human-labeled data.

\boldparagraph{Visualizing Rare Examples}%
We visualize the rarest ground-truth examples from the Waymo Open Dataset as determined by the estimated log probability of every instance. We aggregate the rareness score for every track by taking the mean log probability of the objects from different frames in each track. We then rank the objects by descending average log probability. See \figref{fig:viz_rare_common} for a visualization of the rarest objects in the dataset.

The rarest ground-truth objects include boxes around vehicle parts (protruding ducts, truck loading ramp) and oversized or irregularly shaped vehicles (trucks, flatbed trailers), which match our intuition regarding rare vehicles. Moreover we discover a small number of mislabeled ground-truth instances among the rarest examples. This illustrates that rare example detection is an out-of-distribution detection problem. Intra-class long-tail examples, in one sense, can be defined as in-category, out-of-distribution examples.

\begin{figure}[t!]
    \centering
     \begin{subfigure}[b]{0.5\linewidth}
         \centering
\includegraphics[width=\textwidth]{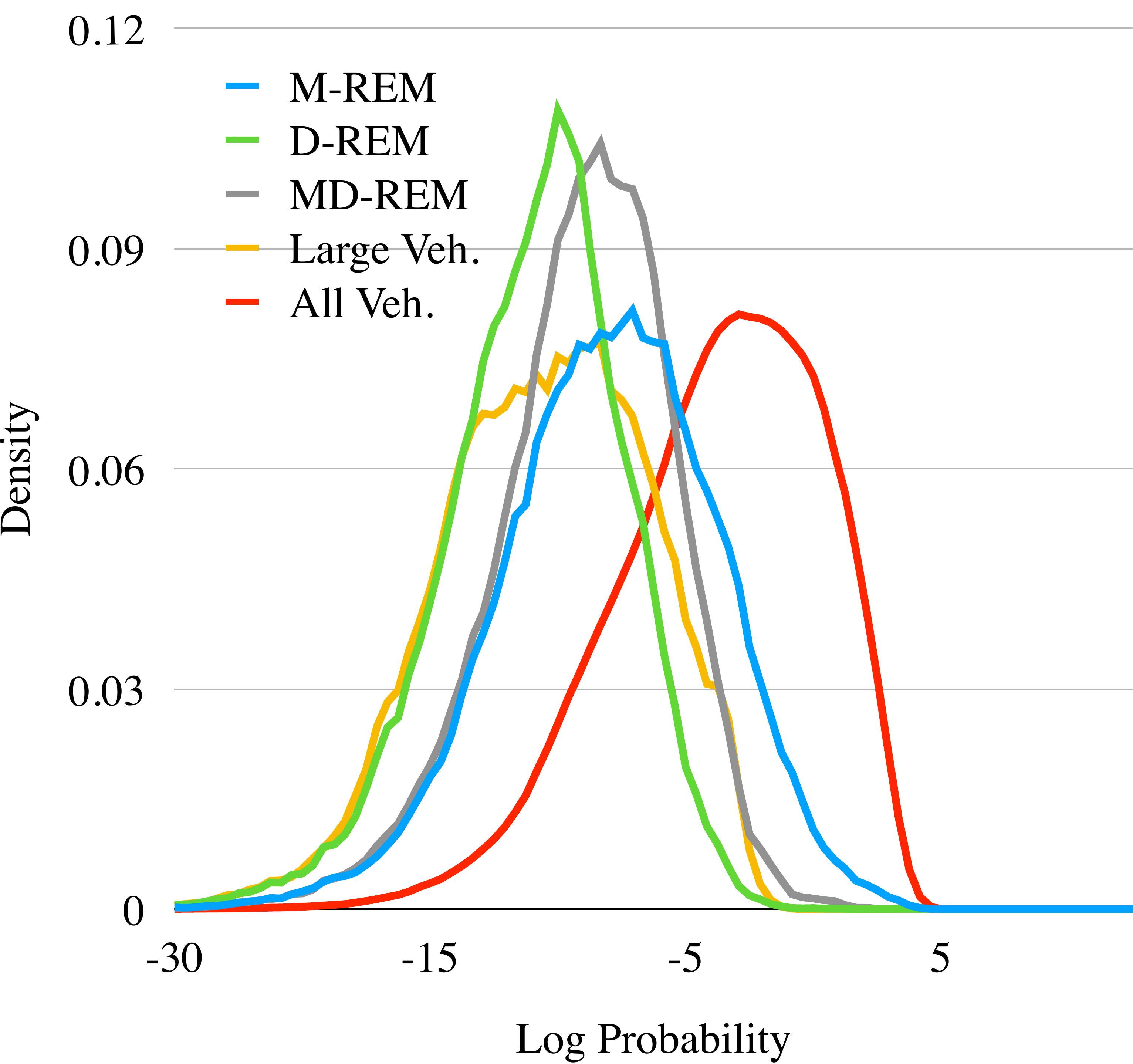}
         \caption{Different vehicle subsets.}
         \label{fig:dist_size}
     \end{subfigure}%
     \hfill
     \begin{subfigure}[b]{0.5\linewidth}
         \centering
         \includegraphics[width=\textwidth]{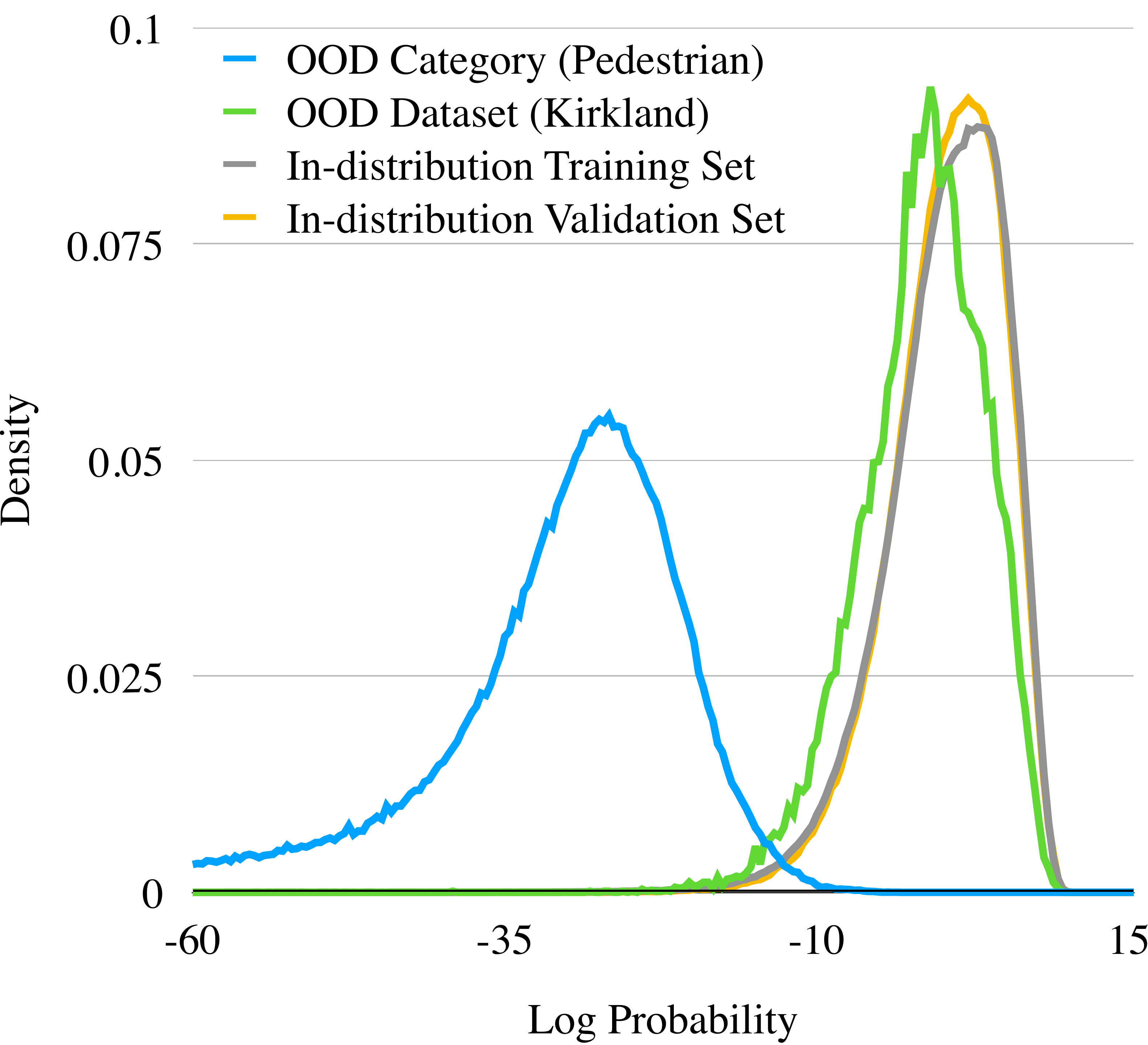}
         \caption{In-domain vs OOD distributions.}
         \label{fig:dist_ood}
     \end{subfigure}
        \caption{Distributional sensitivity of the flow model trained on the vehicle class of the Waymo Open Dataset \cite{sun2020scalability}. (a) Log probability distribution of different vehicle subsets (size subsets and REM mined subsets). (b) Log probability distribution of in-/out-of-distribution examples.}
        \label{fig:dist_sensitivity}
\end{figure}


\boldparagraph{Distributional Sensitivity of the Flow Model}%
In-light of the observation that rare example mining is inherently an out-of-distribution detection problem, we seek to perform a more quantitative analysis of the model's sensitivity to out-of-distribution instances. See \figref{fig:dist_sensitivity} for a detailed breakdown of the analysis. In \figref{fig:dist_size}, we compare the flow model's inferred log probability distributions between vehicle boxes of different sizes. Vehicle size is defined as the max between box length, width and height. We perform a simple partition for all vehicle examples along the size dimension: regular vehicles as $\text{size}\in[3, 7)~\text{(m)}$, and large vehicles as $\text{size}\in[7, \infty)~\text{(m)}$. Our flow model assigns significantly higher overall log probability for the subset of regular-sized vehicles ($96.18\%$ of total), compared to rare subsets such as large vehicles. Note that we leverage vehicle size as an sanity check for the general REM method to distinguish between known rare and common distribution.

Furthermore, we validate that the flow model is effective at detecting out-of-distribution examples (\figref{fig:dist_ood}). The flow model infers almost-identical log probability distributions between the training and validation sets, while assigning lower probabilities to vehicles from an out-of-distribution set (the Kirkland set from the Waymo Open Dataset, collected from a different geographical region with mostly rainy weather condition). Moreover, the model assigns significantly lower probabilities on OOD categories (Pedestrian) if we perform ROI pooling using the pedestrian ground-truth boxes to extract pedestrian feature vectors from the vehicle model and query the log probability distribution against the flow model.  

\subsection{Rare Example Mining for Active Learning}\label{sec:exp_active}
To demonstrate the applicability of the REM approach for targeted improvement of the model's performance in the intra-class long-tail, we utilize track-level REM for active learning, as detailed in \secref{sec:track_rem}.

\boldparagraph{Experiment Setup}%
Our experiment setup is as follows. Following \citet{qi2021offboard}, we perform a random split on the main training set of the Waymo Open Dataset \cite{sun2020scalability} into a $10\%$ fully-labeled source pool, and a remaining $90\%$ as a larger ``unlabeled" pool, from which we withhold ground-truth labels. We first train our main model on the fully-labeled source pool, and perform track-level data mining on the remaining unlabeled pool using various methods, including our proposed data-centric and model-centric REM approaches. For all active learning baseline experiments, we mine for a fixed budget of $1268$ tracks, amounting to $\sim 3\%$ of all remaining tracks.

Our main model consists of a single-frame MVF detector \cite{zhou2020end,qi2021offboard}. While in all baseline experiments we utilize the main model for self-labeling unlabeled tracks in the unlabeled pool, we demonstrate that using a strong offboard 3D auto-labeler \cite{qi2021offboard} trained on the same existing data can further boost the overall performance of our REM approach.

\begin{table}[t]
\begin{floatrow}
\capbtabbox[.48\textwidth]{%
    \centering
    \begin{tabular}{@{}p{12em}l@{}}
    \toprule
    Mining Criteria    & Ratio of Large    \\
    \\ \midrule
    Random Uniform    & 2.60\%           \\
    Ensemble\cite{beluch2018power} & 13.72\%          \\ \midrule
    Model-centric REM & 24.61\%          \\
    Data-centric REM  & 30.60\% \\ 
    Model+Data-centric REM & \textbf{31.86}\% \\ \bottomrule
    \end{tabular}
}{%
    \caption{Composition of mined tracks. We use the ratio of large ($>7$m) objects as a reference for measuring the ratio of rare tracks mined by different approaches. REM is able to mine a higher proportion of rare instances.}
    \label{tab:composition}
}
\capbtabbox[.48\textwidth]{%
    \centering
    {\footnotesize
    \begin{tabular}{@{}p{5em}p{4em}p{2.5em}p{2.5em}l@{}}
    \toprule
    Experiment       & Human Labels & All   & \scalebox{0.8}[1.0]{Regular} & Large \\ \midrule
    \begin{tabular}{@{}l@{}}Fully \\ Supervised \end{tabular} & 100\%;0\%               & 0.895 & 0.900   & 0.602 \\ \midrule
    Ours  & 10\%;3\%              & 0.904 & \textbf{0.904}   & 0.571 \\
    Ours  & 10\%;6\%              & 0.904 & 0.903   & \textbf{0.612} \\
    Our  & 10\%;9\%              & \textbf{0.905} & \textbf{0.904}   & 0.606 \\ \bottomrule
    \end{tabular}
    }
}{%
    \caption{Impact of mining budget on model performance. With a small increase in mining budget (6\%), we (MD-REM++) can match the performance of a fully-supervised model on both ends of the spectrum.}
    \label{tab:ablate_budget}
}
\end{floatrow}
\end{table}

\boldparagraph{Composition of Mined Tracks}%
We first analyze the composition of the mined tracks, in all cases 1268 tracks obtained using various mining approaches (see \tabref{tab:composition}). 

We derive three main findings from the composition analysis: (1) Data-centric REM is able to effectively retrieve known rare subsets, upsampling large objects by as much as $1214\%$. (2) Comparing model-centric REM to ensemble mining method, a simple hard example filtering operator leads to drastically upsampled rare instances, signifying the dichotomy of rare and hard. By using a hard example filter we can significantly increase the ratio of rare examples among mined tracks. (3) Combining model and data-centric REM (by further performing hard example filtering from instanced mined by data-centric REM) further boosts the ratio of large vehicles.

\begin{table}[t]
    {
    \begin{subtable}[t]{.5\linewidth}
        \centering
        \subcaption{\scalebox{0.9}[1.0]{Reference experiments w/o active learning.}}
        \label{tab:ref_no_active}
        {\small
        \adjustbox{max width=\linewidth}{
        \begin{tabular}{@{}p{8.5em}p{3.5em}p{2.5em}p{2.5em}p{2.5em}@{}}
        \toprule
        Experiment  & Human Labels      & All    & \scalebox{0.8}[1.0]{Regular} & Large \\ \midrule
        Partial-supervised    & 10\%;0\% & 0.845 & 0.853 & 0.378    \\
        \scalebox{0.9}[1.0]{Semi-supervised (SL)} & 10\%;0\% & 0.854 & 0.864 & 0.350   \\
        \scalebox{0.9}[1.0]{Semi-supervised (AL)} & 10\%;0\% & 0.902 & 0.910 & 0.419    \\
        Fully-supervised & 100\%;0\% & 0.895 & 0.900 & 0.602    \\ \bottomrule
        \end{tabular}
        }}

        \subcaption{Oracle active learning experiments.}
        \label{tab:oracle_active}
        {\small
        \adjustbox{max width=\linewidth}{
        \begin{tabular}{@{}p{8.5em}p{3.5em}p{2.5em}p{2.5em}p{2.5em}@{}}
        \toprule
        Oracle Hard \cite{shrivastava2016training} & 10\%;3\% & 0.865 & 0.875 & 0.341 \\
        Oracle Size  & 10\%;3\% & 0.869 & 0.875 & 0.583 \\ \bottomrule
        \end{tabular}
        }}
    \end{subtable}%
    \begin{subtable}[t]{.5\linewidth}
        \centering
        \caption{Main active learning experiments.}
        \label{tab:main_active}
        {\small
        \adjustbox{max width=\linewidth}{
        \begin{tabular}{@{}p{8.5em}p{3.5em}p{2.5em}p{2.5em}p{2.5em}@{}}
        \toprule
        Experiment  & Human Labels      & All    & \scalebox{0.8}[1.0]{Regular} & Large \\ \midrule
        Partial-supervised    & 10\%;0\% & 0.845 & 0.853 & 0.378    \\
        Random          & 10\%;3\% & 0.873 & 0.881 & 0.355 \\
        Predict Size    & 10\%;3\% & 0.865 & 0.871 & 0.498 \\
        Ensemble \cite{beluch2018power} & 10\%;3\% & 0.869 & 0.879 & 0.353 \\ \midrule
        Ours (M-REM)    & 10\%;3\% & 0.886 & 0.893 & 0.478 \\
        Ours (D-REM)    & 10\%;3\% & 0.882 & 0.888 & 0.483 \\
        Ours (D-REM++)  & 10\%;3\% & \textbf{0.906} & \textbf{0.913} & 0.533 \\
        \begin{tabular}{@{}l@{}}Ours \\ (MD-REM++)\end{tabular} & 10\%;3\% & 0.904 & 0.909 & \textbf{0.571} \\ \bottomrule
        \end{tabular}
        }}
    \end{subtable}
    \caption{Active learning experiment results. Our method significantly improves model performance across the spectrum, particularly significantly on rare subsets. We denote human label ratio as $(\%s, \%t)$ to indicate the model being trained with $\%s$ of full-labeled run segments, along with $\%t$ of the remaining tracks that is mined and labeled. }
    \label{tab:all_active}
    }
\end{table}

\boldparagraph{Active Learning Experiment}%
We present our active learning experiments in \tabref{tab:all_active}. Results are on vehicles from the Waymo Open Dataset \cite{sun2020scalability}, reported as AP at IoU 0.5. We compute subset metrics on all vehicles (``All"), regular vehicles (``Regular") of size within $3-7$m, large vehicles (``Large") of size $>7$ m. ``Regular" subset is a proxy of the common vehicles, while ``Large" subset is a proxy for rare.

In \tabref{tab:ref_no_active}, we present performances of the single-frame MVF model trained on different compositions of the data. We denote semi-supervised method using self-labeled segments as ``SL" and auto-labeled segments as ``AL". The main observation is that although auto-labeling can significantly improve overall model performance, in particular for common (regular-sized) vehicles, the resulting model performance is significantly weaker for rare subsets, motivating our REM approach.

For the active learning experiments, we first compare two oracle-based approaches (\tabref{tab:oracle_active}) that utilize 100\% ground-truth knowledge for the mining process. ``Oracle Hard" is an error-driven mining method inspired by \cite{shrivastava2016training}, that ranks tracks by $s = \text{IoU(GT, Pred) * Probability\_Score(Pred)}$ to mine tracks which either the base model made a wrong prediction on, or made an inconfident prediction. ``Oracle Size" explicitly mines 3\% of ground-truth tracks whose box size is $>7m$. The main observation is that error-based mining favors difficult examples which do not help improve model performance. Though size-based mining can effectively improve large vehicle performance, it solely improves large vehicles and does not help on regular vehicles.

We then compare across a suite of active learning baselines and our proposed REM methods (\tabref{tab:main_active}). ``Random" mines the tracks via randomized selection, ``Predict Size" mines tracks associated with the largest predicted boxes, and ``Ensemble" mines the tracks with highest ensemble variance (\eqnref{eqn:ens_var}). For our proposed REM methods, we prefix model-centric REM approaches with ``M-", data-centric approaches with ``D-", and a hybrid approach leveraging hard-example filtering on top of data-centric approaches with ``MD-". To further illustrate the importance of a strong offboard auto-labeler, we add auto-labeler to our method, denoting the experiments with ``++".

The active learning experiments show that: (1) Both data-centric and model-centric approaches significantly help to improve performance on the rare subset, and a combination of the two can further boost the long-tail performance, (2) While heuristics based mining methods (``Predict Size") can achieved targeted improvement for large vehicles, it likely fails to capture other degrees of rareness, resulting in lower overall performance.

\section{Ablation studies}
We further study the impact of increasing mining budget on our REM approach (\tabref{tab:ablate_budget}). With a small increase of mining budget (6\%), we can match the performance of a fully-supervised model for both common and rare subsets.

\section{Discussions and Future Work}
In this work, we illustrate the limitations of learned detectors with respect to rare examples in problems with large intra-class variations, such as 3D detection. We propose an active learning approach based on data-centric and model-centric rare example mining which is effective at discovering rare objects in unlabled data. Our active learning approach, combined with a state-of-the-art semi-supervised method can achieve full parity with a fully-supervised model on both common and rare examples, utilizing as little as 16\% of human labels.

A limitation of this study is the scale of the existing datasets for active learning, where data mining beyond the scale of available datasets is limited. Results on a larger dataset will be more informative. Our work shares the same risks and opportunities for the society as other works in 3D detection.

Future work includes extending the REM approach beyond 3D detection, including other topics in self-driving such as trajectory prediction and planning.

\subsubsection{Acknowledgements}
We thank Marshall Tappen, Zhao Chen, Tim Yang, Abhishek Sinh and Luna Yue Huang for helpful discussions, Mingxing Tan for proofreading and constructive feedback, and anonymous reviews for in-depth discussions and feedback.

\clearpage
%
%
\bibliographystyle{splncs04nat}
\bibliography{citations}
\end{document}


\pagestyle{headings}
\mainmatter
\def\ECCVSubNumber{5325}  

\title{Improving the Intra-class Long-tail in 3D Detection via Rare Example Mining \\ ~\\ \large{Supplemental Materials}}

\author{}
\institute{}
\maketitle

\setlength{\abovedisplayskip}{0.25\abovedisplayskip}
\setlength{\belowdisplayskip}{0.25\belowdisplayskip}
\setlength{\abovecaptionskip}{0.5\abovecaptionskip}
\setlength{\belowcaptionskip}{0.25\belowcaptionskip}

\setlength{\textfloatsep}{10pt plus 1.0pt minus 2.0pt}
\setlength{\floatsep}{12pt plus 1.0pt minus 1.0pt}
\newfloatcommand{capbtabbox}{table}[][\FBwidth]
\newfloatcommand{capbalgbox}{algorithm}[][\FBwidth]

\section{Experiment Details}
\label{sec:sup_exp_details}
We provide additional details for our experiments below.

\subsection{Training the Normalizing Flow Model}
We train a continuous normalizing flow model (FFJORD \cite{grathwohl2018ffjord}) for estimating the rareness of detection objects. As detailed in the main text, we train the flow model on the normalized feature vectors after PCA transformation obtained via ROI max-pooling the final feature map of the MVF detector using predicted or ground-truth bounding boxes.

\begin{figure}
     \centering
     \begin{subfigure}[b]{0.3\textwidth}
         \centering
         \includegraphics[width=\textwidth]{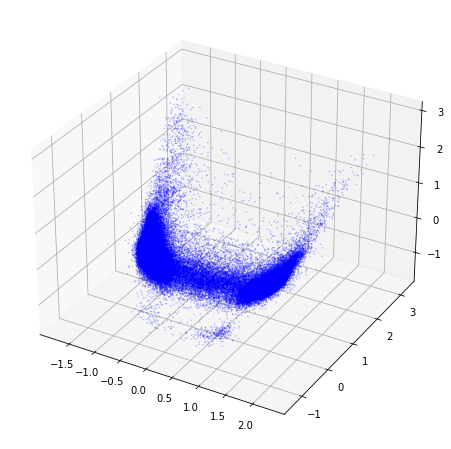}
         \caption{PCA projection of the training embeddings.}
         \label{fig:nf_training_features}
     \end{subfigure}
     \hfill
     \begin{subfigure}[b]{0.3\textwidth}
         \centering
         \includegraphics[width=\textwidth]{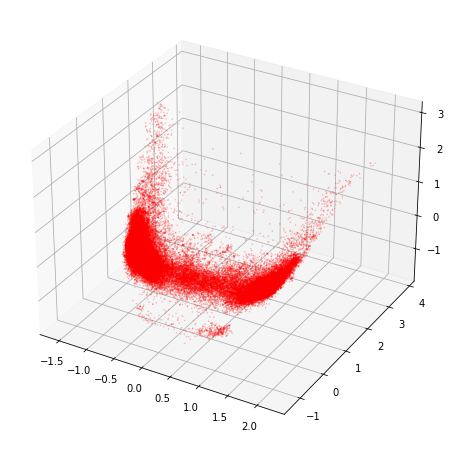}
         \caption{PCA projection of generated embeddings.}
         \label{fig:nf_generated_features}
     \end{subfigure}
     \hfill
     \begin{subfigure}[b]{0.3\textwidth}
         \centering
         \includegraphics[width=\textwidth]{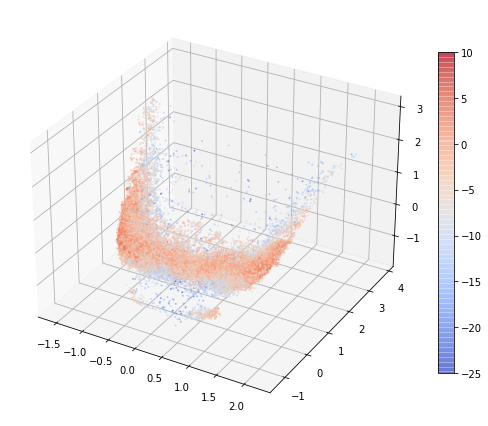}
         \caption{Estimated log probability of embeddings.}
         \label{fig:nf_logp}
     \end{subfigure}
    \caption{Visualization of (a) the embeddings used for training the normalizing flow model, (b) the generated embeddings from the flow model by sampling the learned distribution, that matches the training distribution very well, signifying a good learned estimation of feature densities.(c) a visualization of the estimated log probability of the embeddings from a learned normalizing flow model. The model is able to assign higher log probability to denser features (common examples) and lower log probability to sparser features (rare examples).}
    \label{fig:nf_visualizations}
\end{figure}

The feature vectors are reduced to the dimension of 10 after the PCA transformation. The flow model consists of a stack of 4 consecutive FFJORD bijectors. Each FFJORD bijector consists of 4 hidden layers, each hidden layer consists of 64 units, and uses \texttt{tanh()} activation. The model is trained using an Adam Optimizer, with an initial learning rate of \texttt{1e-4}, learning rate decay every 2400 steps, with a decay rate of 0.98. We train the flow model for a total of 100 epochs. For the base distribution of the flow model, we use a spherical Gaussian distribution with unit variance. See \figref{fig:nf_visualizations} for a visualization of the inferred feature densities using a flow model trained using the procedure above.

\subsection{3D Auto-labeler Implementation}
We adopt the auto-labeling pipeline as outlined in \cite{qi2021offboard}. We use a strong teacher model to serve as the auto-labeler. For the active learning experiments, we train a 5-frame MVF model \cite{zhou2020end} on the same 10\% fully-labeled segments as the single-frame MVF student model. We then use this 5-frame MVF model to extract boxes, create object tracks using \cite{weng20203d}, and further refine the detections using the refiner \cite{qi2021offboard}.

\section{Additional Visualizations of Rare Examples}
\label{sec:supp_viz_rare}

\begin{figure}[h]
    \centering
    \includegraphics[height=0.71\paperheight, width=\textwidth]{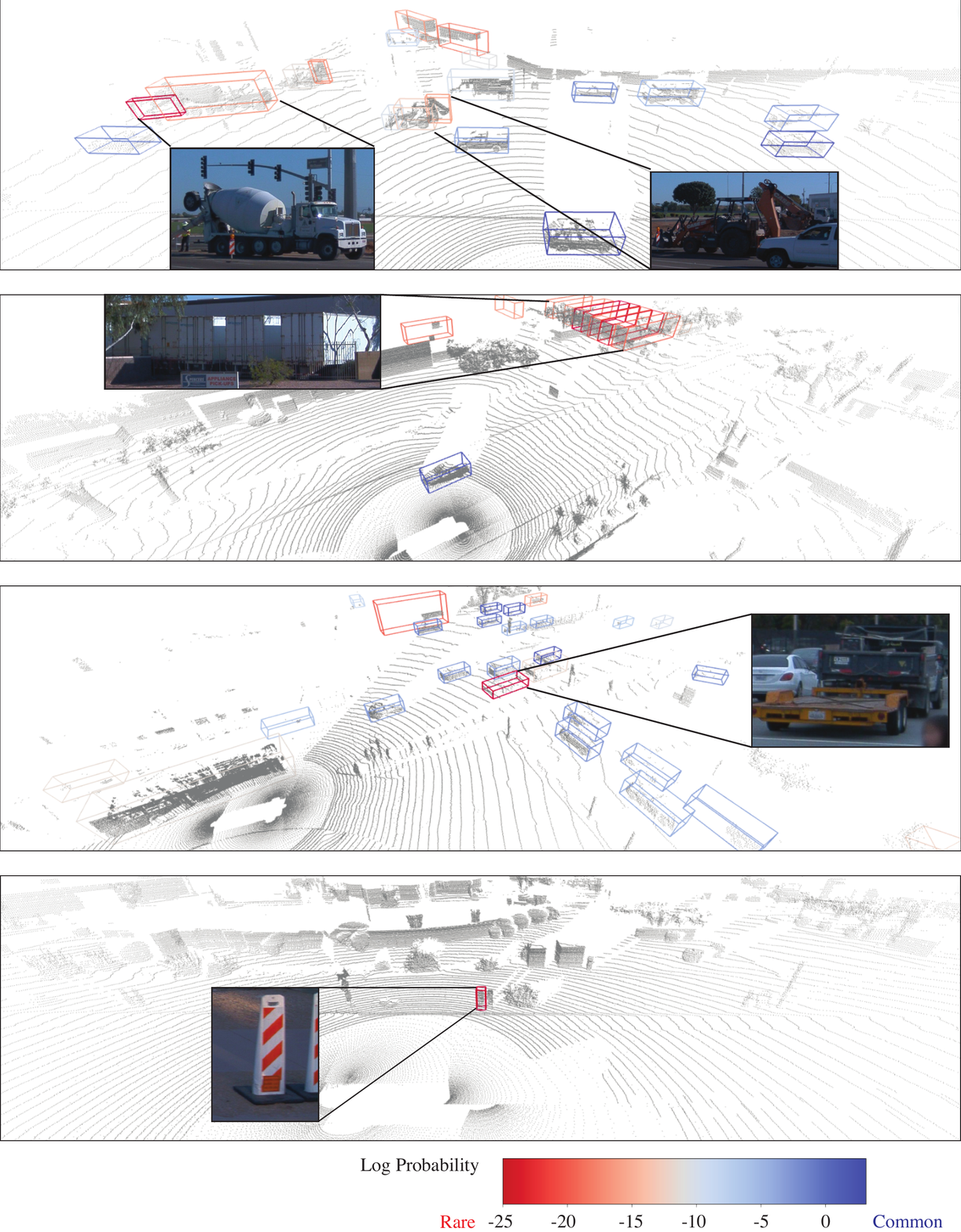}
    \caption{Additional visualizations of rare examples from the Waymo Open Dataset, as inferred by the trained flow model. Warmer colors indicate more rare instances in the scene. The most rare instances in the scene include large vehicles (construction vehicles, trucks), vehicles of irregular geometries such as a flatbed trailer, as well as potentially mislabeled objects such as cones.}
    \label{fig:additional_rare_viz}
\end{figure}

\clearpage

\section{Additional Experimental Evaluation}
\subsection{Pedestrian Rare Example Mining}
We perform an additional experimental evaluation for rare example mining performance on pedestrians to offer more insights into using rare example mining for other object categories. Similar to the main experiments in the paper, we perform this experiment on the Waymo Open Dataset \cite{sun2020scalability}. We use the same MVF \cite{zhou2020end} backbone as the main pedestrian detector and utilize \cite{qi2021offboard} for auto-labeling the unlabeled instances. We mainly compare three sets of experiments. We compare two active learning experiments that train on 10\% of the fully-labeled segments, plus an additional 10\% of mined pedestrian tracks from the rest of the training segments, where the tracks are either mined using our MD-REM++ method or by randomly selecting from the remaining tracks. For our MD-REM++ method, we use a hard example filtering function where the number of points threshold $\tilde{p}=10$ and $d=50$(m). We report on Average Precision (AP) metrics at an IoU threshold of 0.5. Unlike vehicle experiments where we have an intuitive estimate of rareness based on vehicle size, for pedestrian experiment we instead evaluate the model performance on the intra-class long-tail by evaluating the AP performance on the top-5\% rarest ground truth objects (based on inferred log probability). We present our results in \tabref{tab:ped_mining}.

\begin{table}[h]
\centering
\begin{tabular}{@{}llll@{}}
\toprule
Experiment       & Human Labels (\%) & All   & Rare (Top 5\%) \\ \midrule
Fully Supervised & (100\%;0\%)       & 0.757 & 0.111          \\ \midrule
Random           & (10\%;10\%)       & 0.705 & 0.038          \\
Ours (MD-REM++)  & (10\%;10\%)       & \textbf{0.729} & \textbf{0.060}          \\ \bottomrule
\end{tabular}
\caption{Active learning experiments for pedestrians on Waymo Open Dataset.}
\label{tab:ped_mining}
\end{table}

We conclude two main findings from the pedestrian experiment that is consistent with the vehicle experiments:
\begin{enumerate}
    \item Our rare example mining method is able to significantly outperform baselines.
    \item Our flow-based method is able to find challenging instances for the model, as the model performs much worse on rare subsets, compared to the general distribution.
\end{enumerate}
\hfill
\pagebreak
\subsection{Additional Evaluation for Vehicle Rare Example Mining}
We show additional evaluation for the experiments presented in Table 3 (main paper). We present the results below.

\begin{table}[h]
\centering
\begin{tabular}{@{}llllll@{}}
\toprule
Experiment       & Human Labels (\%) & All   & Regular & Large & Rare (Top 5\%) \\ \midrule
Fully Supervised & (100\%;0\%)       & 0.730 & 0.732   & 0.458 & 0.178          \\ \midrule
Ours (MD-REM++)  & (10\%,3\%)        & 0.732 & 0.735   & 0.423 & 0.126          \\
Ours (MD-REM++)  & (10\%,6\%)        & 0.729 & 0.732   & 0.415 & 0.126          \\
Ours (MD-REM++)  & (10\%;9\%)        & 0.730 & 0.732   & 0.443 & 0.152          \\ \bottomrule
\end{tabular}
\caption{Additional evaluations for Table 3 in the main paper. Here we report AP @ IoU 0.7 for these experiments.}
\end{table}

Similar to the pedestrian experiment metrics, we further introduce a Top-5\% rare subset metric. By mining more data via increasing the mining budget using our REM approach, we are able to further close up the gap with the fully labeled model with a much more reduced total labeling cost.

\clearpage

{\small
\bibliographystyle{ieee_fullname_natbib}
\bibliography{citations.bib}
}